\def\year{2022}\relax
\documentclass[letterpaper]{article} 
\usepackage{aaai22}  
\usepackage{times}  
\usepackage{helvet}  
\usepackage{courier}  
\usepackage[hyphens]{url}  
\usepackage{graphicx} 
\usepackage{natbib}  
\usepackage{caption} 
\DeclareCaptionStyle{ruled}{labelfont=normalfont,labelsep=colon,strut=off} 
\usepackage{algorithm}
\usepackage{algorithmic}

\usepackage{newfloat}
\usepackage{xcolor}
\usepackage{listings}
\floatstyle{ruled}
\newfloat{listing}{tb}{lst}{}
\floatname{listing}{Listing}
\copyrighttext{Presented at the AI-HRI Symposium at AAAI Fall Symposium Series (FSS) 2022}
\title{Towards Cognitive Robots That People Accept in Their Home}

\author {
    Nina Moorman, 
    Erin Hedlund-Botti, 
    Matthew Gombolay 
}
\affiliations {
    Georgia Institute of Technology \\
    School of Interactive Computing \\
    Atlanta, Georgia, USA \\
   ninamoorman@gatech.edu, erin.botti@gatech.edu, matthew.gombolay@cc.gatech.edu
}

\begin{document}
\maketitle

\begin{abstract}
It is intractable for assistive robots to have all functionalities pre-programmed prior to deployment. Rather, it is more realistic for robots to perform supplemental, on-site learning about user's needs and preferences, and  particularities of the environment. This additional learning is especially helpful for care robots that assist with individualized caregiver activities in residential or assisted living facilities. As each care receiver has unique needs and those needs may change over time, robots require the ability to adapt and learn on-site. In this work, we propose the study design to investigate the impacts on end-users of observing robot learning. We will assess user attitudes towards robots that conduct some learning in the home as compared to a baseline condition where the robot is delivered fully capable. We will additionally compare different modes of learning to determine whether some are more likely to instill trust.
\end{abstract}


\section{Introduction}
\label{sec:introduction}
Care robots perform a wide variety of assistive tasks that could benefit from some degree of in situ learning. This supplemental learning would enable the robot to adapt to its environment and users. It would additionally offer caregivers and care receivers the option of being directly involved in the robot's learning, enabling end-users to specify preferences or teach the robot new tasks directly. This customization is important for ensuring effective, individualized care. Though at-home learning is already in use for some care robots, we lack an understanding of how observing this learning affects user trust \cite{irobot_irobot_nodate, moxi, paro}. This work aims to develop a better understanding of how users will respond to observable robot learning. 

We propose a series of human-subjects experiments that evaluate user perceptions of learning robots using both surveys and behavioral metrics. 
We choose four learning methods to understand how user perception of a learning robot may vary depending on the degree of user involvement in the learning process. We capture low involvement learning with a reinforcement learning (RL) condition, and high involvement learning with an learning from demonstration (LfD) condition using kinesthetic demonstrations. Our control baseline is a download condition where the robot downloads tasks from the cloud. Finally, we include the Training an Agent Manually via Evaluative Reinforcement (TAMER) \cite{knox2009interactively} condition which is a middle ground between RL an LfD. 
We will compare robot perception in each of the learning conditions for in person and remote participants, and for both the caregiver and general population. 

\begin{figure}[t]
    \centering
    \includegraphics[scale=0.45]{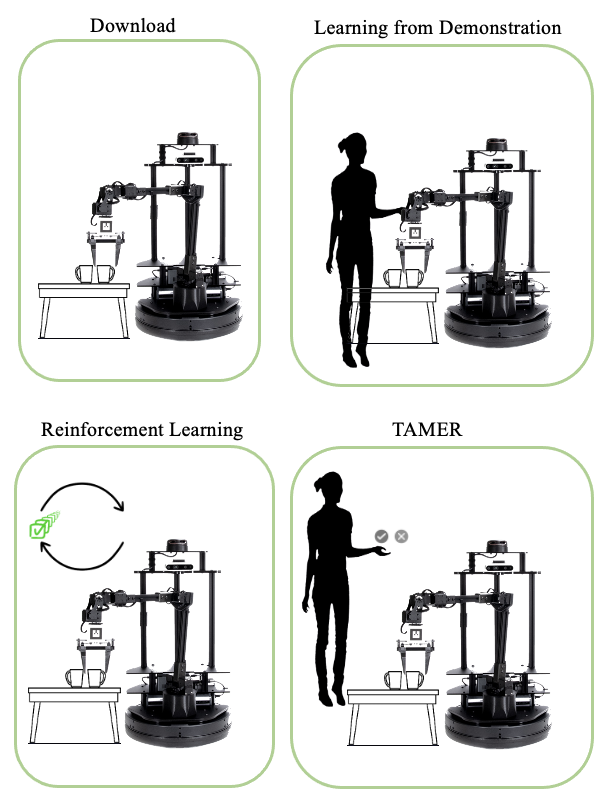}
  \caption{This figure shows the four learning conditions.}
    \label{fig:agent_types}
\end{figure}

Informed by the findings we then wish to determine the techniques that are most effective in repairing trust in a learning robot. As learning agents may not learn quickly and may fail often, it is important to evaluate the best method of trust repair for embodied, learning agents \cite{3374839}. Finally, we aim to develop guidelines to inform the design of care robotic systems that learn and operate in residential or nursing-home environments. In our work we propose the following:

\begin{enumerate}
    \item Study the difference between user perceptions of a fully pre-engineered robot compared to a learning robot.	
    \item Investigate how to best perform trust repair with respect to embodied robots that learn in the home.
    \item Develop guidelines to inform the design of assistive robotic systems deployed in residential environments.
\end{enumerate}

\section{Related Works}
\label{sec:relatedworks}

\subsubsection{Care Robots -}
Care robots are defined by their function to support caregivers and/or care receivers \cite{van_wynsberghe_designing_2013}. These robots can operate in residential environments where they perform a variety of assistive tasks and promote extended independent living \cite{johnson_socially_2014,sabanovic_robot_2015, fiorini_assistive_2021, lee_reframing_2018}. Care robot roles generally fall under physical assistance or medical assistance. Physical assistance includes tasks such as navigation, fall-prevention, object manipulation, and household chores \cite{moxi, fischinger_hobbit_2016, swisslog_healthcare_autonomous_nodate, kittmann_let_2015, kostavelis_ramcip_2016, miseikis_lio-personal_2020}. Medical assistance includes tasks such as health monitoring, medicine delivery, and the exertion of a social presence for coaching or social interaction \cite{vitanza_assistive_2019, coradeschi_giraffplus_2013, umbrico_holistic_2020, nao, pepper, martinez-martin_pharos_2019}.

On-site learning affords the robot an opportunity to observe and adapt to individual user needs and preferences, which may be beneficial for both physical and medical assistance tasks. In this work, we seek to understand how observing various forms of robot learning may impact user perceptions of care robots. 

\subsubsection{Acceptance and Trust -}
Robot acceptance depends not only on the benefits the robot can bestow upon the user, but also on the user's perception of and attitudes towards the robot \cite{cesta_psychological_2007}. One of the most important attitudes with respect to acceptability is user trust \cite{yagoda_you_2012, langer_trust_2019}. Trust is defined as a user’s attitude that the agent will help them achieve a goal, specifically in a situation of uncertainty, or vulnerability \cite{kohn_measurement_2021, ullman_what_2018}. 

Prior work in human-automation (HA) trust has categorized trust based on the extent of interaction with the user into dispositional, situational, and learned trust \cite{hoff_trust_2015}. These represent baseline trust in automation \cite{merritt_i_2013}, trust with respect to a particular interaction \cite{jian_foundations_2000}, and trust developed though a series of interactions \cite{de_visser_towards_2020}. 

Another approach for investigating HR trust is to study trust dependent on robot-specific factors, such as performance \cite{hedlund_effects_2021}. In our work, we keep performance consistent between conditions to isolate the impact of observing the learning process. However, there may still be differences in perceived robot ability due to the observation of different forms of learning; as such, we will evaluate performance-based trust. 

In human robot interactions, anthropomorphism (i.e., the degree to which a robotic agent demonstrates human-like characteristics) has been shown to affect trust \cite{3374839}. Thus, we will hold the robot's embodiment constant throughout the study and subjectively measure subjects' perception of the robot's anthropomorphism to control for this effect. The robot we employ is the Movo Beta robot. Finally, in our work, we conduct both in-person and remote studies to account for the impact of the robot's physical presence on user trust. 




\section{Methodology}
\label{sec:methods}
This section details the domain, experimental populations, learning conditions, hypotheses, metrics, and study design. 

\subsection{Domain}
\label{sec:domain}
\begin{figure}[ht]
    \centering
    \includegraphics[width=1\columnwidth]{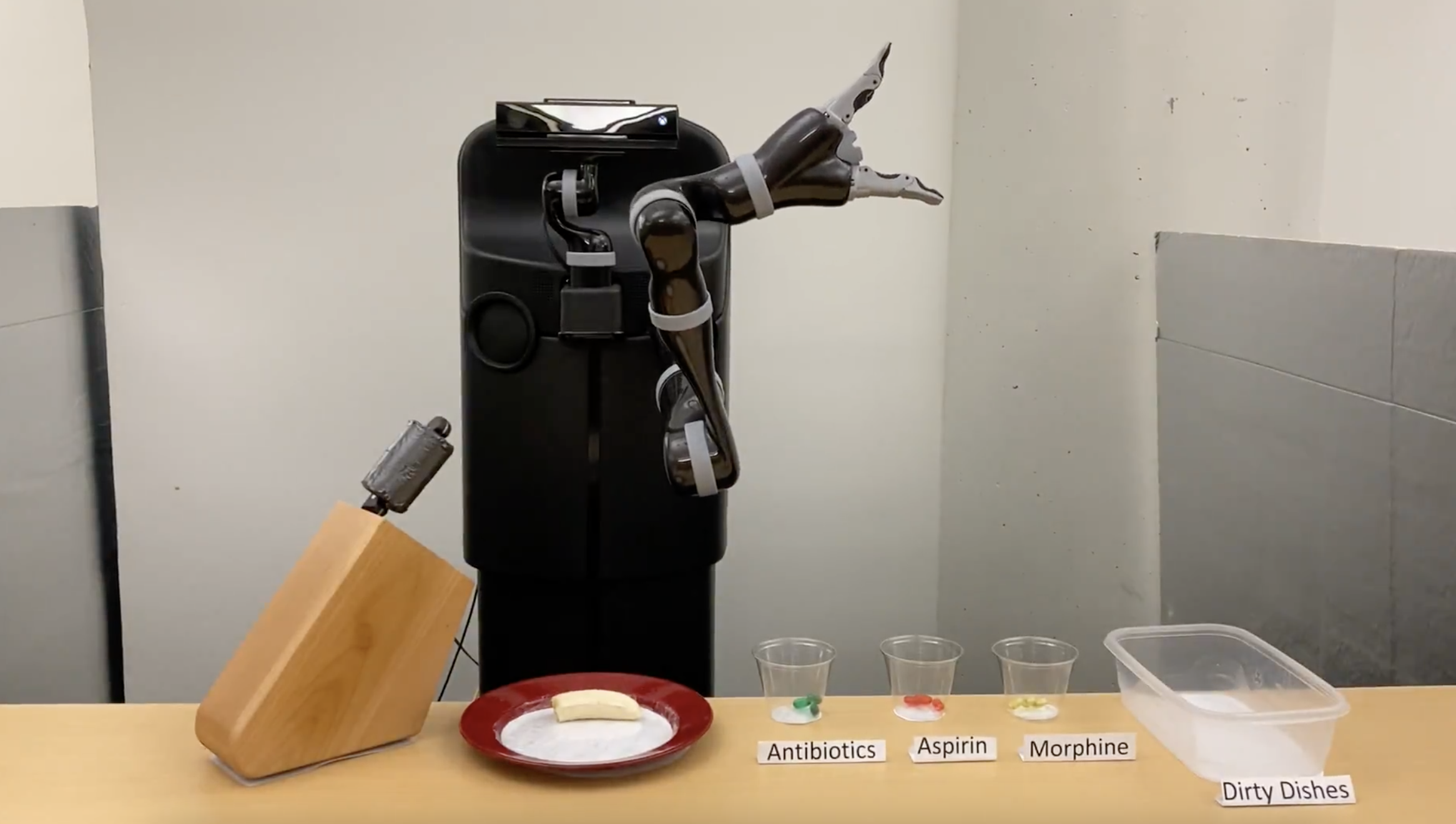}
  \caption{This figure shows the testing phase study design.}
    \label{fig:study_setup}
\end{figure}

The plate-making domain we choose is depicted in Figure \ref{fig:study_setup}. This task involves picking up a knife, cutting a banana that has already been placed on a plate in half, then picking the correct medicine, and pouring it on to the plate. We choose this task as it is a combination of both a cognitive preparation task (recipe-following) and the manipulation task (cutting). We isolate participant perceptions of the robot reported in each of these sub-tasks. 

In the training phase the knife used will be a small plastic knife, and the medicine dispensed will be composed of different types of vitamins. We make the testing phase riskier by changing the knife to a large, sharp, metal knife, and changing the medicine to prescription grade medications labeled as morphine, aspirin, and antibiotics. Recall from our definition of trust that we must evaluate trust in a situation of uncertainty, or vulnerability \cite{kohn_measurement_2021, ullman_what_2018}. As a result, we raise the stakes in the test task to ensure that participants consider their risk tolerance when evaluating the robot.

\subsection{Experiment Populations} \label{sec:humanexperiment}
Virtual studies may not fully represent the impact of the robot's physical presence on trust. However, conducting our study in person would impose a high health risk and a potentially impractical transportation burden on our target population of caregivers. As such, we propose to conduct three human-subjects experiments.

\indent \textit{Study 1: Remote, General Population} We invite members accessible to us at a metropolitan university campus to take our study virtually. 

\indent \textit{Study 2: In Person, General Population} We invite members of the local population to take our study in the laboratory. In Study 2, the test phase will be conducted live, in person.

\indent \textit{Study 3: Remote, Caregivers} We invite our target population of caregivers and elder care nurses in our state to take our study virtually.

By comparing the results of Study 1 and Study 2, we determine the influence of embodied robots on user trust. Additionally, by comparing the results of Study 1 and Study 3, we determine the difference in trust between caregivers and the general population.

\subsection{Learning Conditions} 
\label{sec:conditions}
In this section, we describe the training video for each of the four learning conditions. For consistency, the training in all studies (virtual and in person) is recorded and viewed in a video format. The method by which the robot learns depends upon the learning condition. We aim to investigate how people may feel differently about the robot depending on the level of user-involvement in the robot learning. Thus, the learning conditions reflect different levels of user involvement.

\textit{Download: }In the download condition, the participant observes the robot download the task knowledge from ``the cloud.'' This serves as the control condition, as no learning is observed by the user. 

\textit{RL: }In the RL condition, the robot demonstrates trial-and-error learning, iteratively learning sub-tasks of the overall task until the task is completed. No explicit reward function will be explained to the participant, and we intentionally describe stages of the learning vaguely, using terms such as start, middle, and end of training rather than providing training duration or iteration count.

\textit{LfD: }In the LfD condition, the same trial-and-error learning is observed; however, we intersperse videos of a human teaching the robot the sub-tasks prior to improvement in performance on these sub-tasks.

\textit{TAMER: }The TAMER condition is a middle ground between LfD and RL, where the robot attempts the task on its own and the user provides binary feedback throughout the learning process to shape the robot's behavior. This feedback will be shown through a graphic of a remote control with green and red buttons pressed during training to convey positive and negative feedback.

\subsection{Hypotheses}
\label{sec:hypotheses}

\textbf{Hypothesis 1} \textit{We hypothesize that participants will trust and adopt robots whose learning they have observed more than robots whose learning was not observable.} We postulate that participants will feel that they understand and can relate more to a robot that learns visibly (RL, LfD, TAMER conditions) than a robot whose learning is not observed (download condition), leading to differences in trust.

\textbf{Hypothesis 2} \textit{We hypothesize that participants will trust and adopt a robot that learns via LfD more than the other learning conditions.} LfD has been shown to be the most intuitive method of interacting with robotic agents, thus, we hypothesize that participants will demonstrate higher trust in and adoption of robots that employ this form of robot learning \cite{7451754, Akgun2011RobotLF}.

\textbf{Hypothesis 3} \textit{We hypothesize that participants will trust a robots less if it is physically present as compared to a remote robot.} Given that in-person participants (Study 2) experience the risk of embodied learning in person, we posit that these participants will trust the robot less than participants for which the task's risk is virtual (Study 1).

\textbf{Hypothesis 4} \textit{We hypothesize that the caregiver population will report lower trust in the robot than the general population.} We posit that the caregiver population (Study 3) will find the risk of robot error on physical manipulation and medicine dispensing tasks performed for care receivers to be more tangible and severe than the general population (Study 2), resulting in lower robot trust.

\subsection{Metrics}
\label{sec:metrics}
We will evaluate a user's dispositional trust, situational trust, and performance-based trust through surveys \cite{merritt_i_2013, jian_foundations_2000, malle_multidimensional_2021}. We will also study user trust using behavioral measures (i.e., in terms of reliance on and compliance with the robot) by determining the average intervention rate of participants while observing the robot's behavior on the test task.

\textit{Pre-Study Questionnaire -} In the pre-study questionnaire, we will collect demographic information including participants' education \cite{raub_correlates_nodate}, computer science and robotics prior experience \cite{raub_correlates_nodate}, personality \cite{donnellan_mini-ipip_2006}, field of occupation \cite{noauthor_college_nodate}, and dispositional trust \cite{merritt_i_2013}. We additionally collect users' perception of the robot's anthropomorphism \cite{bartneck_measurement_2009}, usability, and acceptability \cite{belanche_integrating_2012}.

\textit{Post-Trial Questionnaire -}
After each testing trial we will ask participants to rate the degree to which they feel the robot accomplished the task, as well as the degree to which they feel the robot behaves safely \cite{bartneck_measurement_2009}.

\textit{Post-Study Questionnaire -}
In the post-survey questionnaire, we will collect participants' perceived situational trust \cite{jian_foundations_2000}, performance-based trust \cite{malle_multidimensional_2021}, and how risky they perceived the task to be \cite{fischhoff_how_1978}. We will also ask two ad hoc questions. First, we will ask what tasks -- from a list of hand-crafted tasks both in and outside of the distribution of tasks observed in the study -- participants would trust the robot to do. This question measures the extent of user adoption. Secondly, we will ask an open-ended question about the participant’s understanding of the robot’s learning and perception of robot competence. This second question is to collect qualitative information about user assumptions regarding the robot's learning and competence.

\subsection{Procedure}
\label{sec:procedure}
Participants first read and sign the consent form, after which they are assigned a unique and random user ID. Next, participants will watch the unboxing video in which the robotic agent introduces itself and demonstrates its range of mobility and degrees of freedom. After watching the unboxing video, participants will fill out the pre-study questionnaire. 

\begin{figure}[ht]
    \centering
    \includegraphics[width=1\columnwidth]{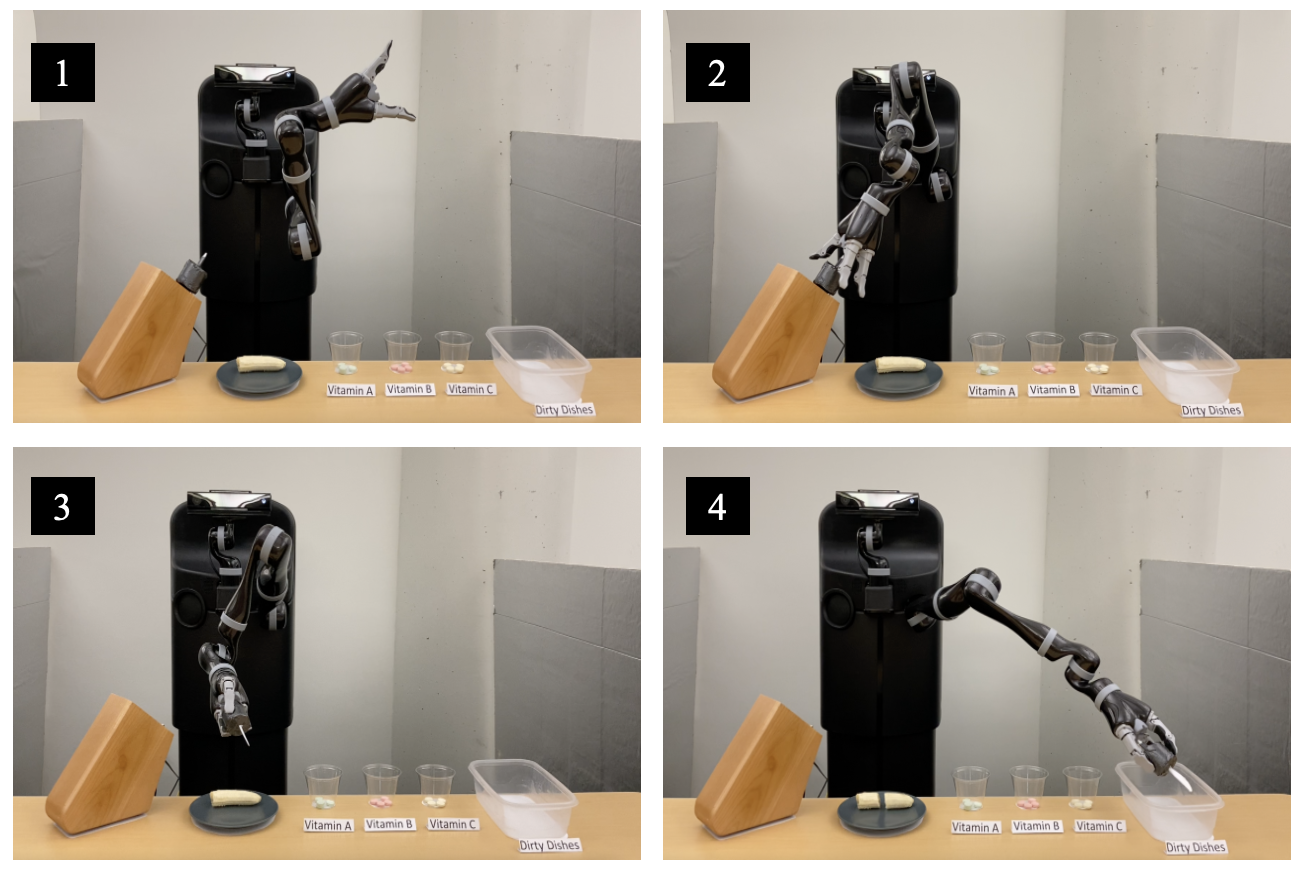}
  \caption{This figure shows a sample training trajectory for the cutting task.}
    \label{fig:training_flow}
\end{figure}

Then, participants will go through the training phase where they will observe the robot learning to perform the training task, as seen in Figure \ref{fig:training_flow}. Ours will be a between-subjects experiment where participants experience one of the four learning conditions. Therefore, in the training phase, participants will watch their condition's unique training video. The only difference between learning conditions will be the type of robot learning observed. After this training phase all participants will observe the same final performance video, and final performance will be held constant between learning conditions. 

Next is the testing phase, where participants will observe nine testing trials where the robot states its goal and attempts to accomplish this goal, with an overall success rate of 80\%. During each testing trial, the participant will be instructed to interrupt the robot by clicking a red stop button if they feel that the robot might be acting in an unsafe manner or if they feel unsafe or uncomfortable, or if they feel that the robot will fail, or will not accomplish the goal of that trial. The interruption data collected here serves to assess reliance. This binary interruption metric, along with the duration of time observing the agent prior to interruption, help to support our findings on trust. After each testing trial, participants will fill out the post-trial questionnaire where we will ask them to rate the degree to which they trust the robot to act safely and the degree to which they believe the robot will accomplish the task. The testing iterations will be shown in person for Study 2 and shown as recorded videos for Study 1 and 3. After the testing phase, the participants will complete the post-study questionnaire.

\subsection{Proposed Analysis}
For the data collected in the Post-Study Questionnaire that passes parametric assumptions of normality and homoscedasticity, we will compare each metric across conditions/populations using ANOVAs with Tukey post-hoc corrections. If the data does not pass these assumptions, we will use non-parametric tests such as the Kruskal-Wallis test with Wilcoxon pairwise tests and Bonferonni post-hoc correction. We will additionally analyze each of the metrics in the Post-Trial Questionnaire using a repeated measures analysis to distinguish between user perception of the robot in the knife and medicine sub-tasks. The information collected from the Pre-Study Questionnaire will be used to determine any potential confounds in the analysis. For Study 1, we plan to run 15 participants in each learning condition, 60 total, with a power of $.8$ and $\alpha=.05$; a power analysis on these values yields a large effect size of $.44$. If we run 60 participants for both in person and remote conditions (Study 1 and Study 2), with a power of $.8$ and $\alpha=.05$, the power analysis yields a medium effect size of $.26$. We aim to recruit at least 12 caregiver participants for Study 3. Given the smaller sample size of the target population, we propose to analyze trends between the general population and caregiver population.

\section{Limitations}
\label{sec:limitations}
One limitation of our work is that in Study 1 and 3 the robot is not physically present with the participant. We are thus investigating user perception of the robot based upon the users' experience watching videos and imagining the robot learning in their home. We aim to quantify the impact of this limitation with Study 2. 

Another limitation of our work is that we constrain our definition of caregiver to nurses employed in assisted living facilities for ease of recruitment in our first investigation. In future work, we plan to increase the breadth of caregiver recruitment to include caregivers who are not nurses (e.g., adult children of parents receiving care). 


\section{Future Work}
\label{sec:futurework}
In future work we will conduct the studies proposed in this paper. Based on these results, we will design a new study (i.e., Study 4) in which we compare various trust repair techniques, applied to a robot that employs the highest effect learning method from Studies 1-3. In Study 4, we propose to evaluate the following three forms of trust repair established in prior work \cite{de_visser_towards_2020, baker_toward_2018, robinette_timing_2015, kim_repairing_2013}.

\begin{enumerate}
    \item An apology provided directly after the trust violation.
    \item Transparency of robot learning, provided as a high-level narration of what is learned.
    \item An explanation of what caused the error, without acknowledging fault, provided after the trust violation.
\end{enumerate}

\section{Conclusion}
\label{sec:conclusion}
We propose a series of human-subject experiments to assess user attitudes towards the concept of embodied care robots that learn in the home, as compared to robots that are delivered fully capable. We investigate the impact of the robot's physical presence on a user's perception of the robot, as well as the differences in robot perception between the general population and caregivers. Based on the findings of our work, we propose to develop guidelines that inform the design of care robots deployed in the home. Finally, we propose to investigate how we can best calibrate trust in embodied learning robots.

\section{Acknowledgments}
This work was supported by NSF IIS-2112633.

\bibliography{aaai22}

\begin{thebibliography}{43}
\providecommand{\natexlab}[1]{#1}

\bibitem[{ACT(2015)}]{noauthor_college_nodate}
ACT. 2015.
\newblock College {Majors} and {Occupational} {Choices}.

\bibitem[{Akgun and Subramanian(2011)}]{Akgun2011RobotLF}
Akgun, B.; and Subramanian, K. 2011.
\newblock Robot Learning from Demonstration : Kinesthetic Teaching vs .
  Teleoperation.

\bibitem[{Baker et~al.(2018)Baker, Phillips, Ullman, and
  Keebler}]{baker_toward_2018}
Baker, A.~L.; Phillips, E.~K.; Ullman, D.; and Keebler, J.~R. 2018.
\newblock Toward an {Understanding} of {Trust} {Repair} in {Human}-{Robot}
  {Interaction}: {Current} {Research} and {Future} {Directions}.
\newblock \emph{ACM Transactions on Interactive Intelligent Systems}, 8(4):
  30:1--30:30.

\bibitem[{Bartneck et~al.(2009)Bartneck, Kulić, Croft, and
  Zoghbi}]{bartneck_measurement_2009}
Bartneck, C.; Kulić, D.; Croft, E.; and Zoghbi, S. 2009.
\newblock Measurement {Instruments} for the {Anthropomorphism}, {Animacy},
  {Likeability}, {Perceived} {Intelligence}, and {Perceived} {Safety} of
  {Robots}.
\newblock \emph{International Journal of Social Robotics}, 1(1): 71--81.

\bibitem[{Belanche, Casaló, and Flavián(2012)}]{belanche_integrating_2012}
Belanche, D.; Casaló, L.~V.; and Flavián, C. 2012.
\newblock Integrating trust and personal values into the {Technology}
  {Acceptance} {Model}: {The} case of e-government services adoption.
\newblock \emph{Cuadernos de Economía y Dirección de la Empresa}, 15(4):
  192--204.

\bibitem[{Cesta et~al.(2007)Cesta, Cortellessa, Giuliani, Pecora, Scopelliti,
  and Tiberio}]{cesta_psychological_2007}
Cesta, A.; Cortellessa, G.; Giuliani, M.~V.; Pecora, F.; Scopelliti, M.; and
  Tiberio, L. 2007.
\newblock Psychological {Implications} of {Domestic} {Assistive} {Technology}
  for the {Elderly}.
\newblock \emph{PsychNology Journal}, 5: 229--252.

\bibitem[{Coradeschi et~al.(2013)Coradeschi, Cesta, Cortellessa, Coraci,
  Gonzalez, Karlsson, Furfari, Loutfi, Orlandini, Palumbo, Pecora, von Rump,
  Štimec, Ullberg, and Ötslund}]{coradeschi_giraffplus_2013}
Coradeschi, S.; Cesta, A.; Cortellessa, G.; Coraci, L.; Gonzalez, J.; Karlsson,
  L.; Furfari, F.; Loutfi, A.; Orlandini, A.; Palumbo, F.; Pecora, F.; von
  Rump, S.; Štimec, A.; Ullberg, J.; and Ötslund, B. 2013.
\newblock {GiraffPlus}: {Combining} social interaction and long term monitoring
  for promoting independent living.
\newblock In \emph{2013 6th {International} {Conference} on {Human} {System}
  {Interactions} ({HSI})}, 578--585.
\newblock ISSN: 2158-2254.

\bibitem[{de~Visser et~al.(2020)de~Visser, Peeters, Jung, Kohn, Shaw, Pak, and
  Neerincx}]{de_visser_towards_2020}
de~Visser, E.~J.; Peeters, M. M.~M.; Jung, M.~F.; Kohn, S.; Shaw, T.~H.; Pak,
  R.; and Neerincx, M.~A. 2020.
\newblock Towards a {Theory} of {Longitudinal} {Trust} {Calibration} in
  {Human}–{Robot} {Teams}.
\newblock \emph{International Journal of Social Robotics}, 12(2): 459--478.

\bibitem[{{Diligent Robotics}(2017)}]{moxi}
{Diligent Robotics}. 2017.
\newblock {Moxi}.
\newblock Https://www.diligentrobots.com/moxi.

\bibitem[{Donnellan et~al.(2006)Donnellan, Oswald, Baird, and
  Lucas}]{donnellan_mini-ipip_2006}
Donnellan, M.; Oswald, F.; Baird, B.; and Lucas, R. 2006.
\newblock The {Mini}-{IPIP} {Scales}: {Tiny}-yet-{Effective} {Measures} of the
  {Big} {Five} {Factors} of {Personality}.
\newblock \emph{Psychological assessment}, 18: 192--203.

\bibitem[{Fiorini et~al.(2021)Fiorini, De~Mul, Fabbricotti, Limosani, Vitanza,
  D’Onofrio, Tsui, Sancarlo, Giuliani, Greco, Guiot, Senges, and
  Cavallo}]{fiorini_assistive_2021}
Fiorini, L.; De~Mul, M.; Fabbricotti, I.; Limosani, R.; Vitanza, A.;
  D’Onofrio, G.; Tsui, M.; Sancarlo, D.; Giuliani, F.; Greco, A.; Guiot, D.;
  Senges, E.; and Cavallo, F. 2021.
\newblock Assistive robots to improve the independent living of older persons:
  results from a needs study.
\newblock \emph{Disability and Rehabilitation: Assistive Technology}, 16(1):
  92--102.
\newblock Publisher: Taylor \& Francis \_eprint:
  https://doi.org/10.1080/17483107.2019.1642392.

\bibitem[{Fischer et~al.(2016)Fischer, Kirstein, Jensen, Krüger, Kukliński,
  aus~der Wieschen, and Savarimuthu}]{7451754}
Fischer, K.; Kirstein, F.; Jensen, L.~C.; Krüger, N.; Kukliński, K.; aus~der
  Wieschen, M.~V.; and Savarimuthu, T.~R. 2016.
\newblock A comparison of types of robot control for programming by
  Demonstration.
\newblock In \emph{2016 11th ACM/IEEE International Conference on Human-Robot
  Interaction (HRI)}, 213--220.

\bibitem[{Fischhoff et~al.(1978)Fischhoff, Slovic, Lichtenstein, Read, and
  Combs}]{fischhoff_how_1978}
Fischhoff, B.; Slovic, P.; Lichtenstein, S.; Read, S.; and Combs, B. 1978.
\newblock How {Safe} {Is} {Safe} {Enough}? {A} {Psychometric} {Study} of
  {Attitudes} {Toward} {Technological} {Risks} and {Benefits}.
\newblock \emph{Policy Sciences}, 9: 127--152.

\bibitem[{Fischinger et~al.(2016)Fischinger, Einramhof, Papoutsakis,
  Wohlkinger, Mayer, Panek, Hofmann, Koertner, Weiss, Argyros, and
  Vincze}]{fischinger_hobbit_2016}
Fischinger, D.; Einramhof, P.; Papoutsakis, K.; Wohlkinger, W.; Mayer, P.;
  Panek, P.; Hofmann, S.; Koertner, T.; Weiss, A.; Argyros, A.; and Vincze, M.
  2016.
\newblock Hobbit, a care robot supporting independent living at home: {First}
  prototype and lessons learned.
\newblock \emph{Robotics and Autonomous Systems}, 75: 60--78.

\bibitem[{Healthcare(2013)}]{swisslog_healthcare_autonomous_nodate}
Healthcare, S. 2013.
\newblock Autonomous {Service} {Robot} {For} {Hospitals} - {Savioke} {Relay}.

\bibitem[{Hedlund, Johnson, and Gombolay(2021)}]{hedlund_effects_2021}
Hedlund, E.; Johnson, M.; and Gombolay, M. 2021.
\newblock The {Effects} of a {Robot}'s {Performance} on {Human} {Teachers} for
  {Learning} from {Demonstration} {Tasks}.
\newblock In \emph{Proceedings of the 2021 {ACM}/{IEEE} {International}
  {Conference} on {Human}-{Robot} {Interaction}}, {HRI} '21, 207--215. New
  York, NY, USA: Association for Computing Machinery.
\newblock ISBN 978-1-4503-8289-2.

\bibitem[{Hoff and Bashir(2015)}]{hoff_trust_2015}
Hoff, K.; and Bashir, M. 2015.
\newblock Trust in {Automation}: {Integrating} {Empirical} {Evidence} on
  {Factors} {That} {Influence} {Trust}.

\bibitem[{iRobot(2002)}]{irobot_irobot_nodate}
iRobot. 2002.
\newblock {iRobot}®: {Robot} {Vacuum} and {Mop}.

\bibitem[{Jian, Bisantz, and Drury(2000)}]{jian_foundations_2000}
Jian, J.-Y.; Bisantz, A.; and Drury, C. 2000.
\newblock Foundations for an {Empirically} {Determined} {Scale} of {Trust} in
  {Automated} {Systems}.
\newblock \emph{International Journal of Cognitive Ergonomics}, 4: 53--71.

\bibitem[{Johnson et~al.(2014)Johnson, Cuijpers, Juola, Torta, Simonov,
  Frisiello, Bazzani, Yan, Weber, Wermter, Meins, Oberzaucher, Panek,
  Edelmayer, Mayer, and Beck}]{johnson_socially_2014}
Johnson, D.~O.; Cuijpers, R.~H.; Juola, J.~F.; Torta, E.; Simonov, M.;
  Frisiello, A.; Bazzani, M.; Yan, W.; Weber, C.; Wermter, S.; Meins, N.;
  Oberzaucher, J.; Panek, P.; Edelmayer, G.; Mayer, P.; and Beck, C. 2014.
\newblock Socially {Assistive} {Robots}: {A} {Comprehensive} {Approach} to
  {Extending} {Independent} {Living}.
\newblock \emph{International Journal of Social Robotics}, 6(2): 195--211.

\bibitem[{Kim et~al.(2013)Kim, Cooper, Dirks, and Ferrin}]{kim_repairing_2013}
Kim, P.~H.; Cooper, C.~D.; Dirks, K.~T.; and Ferrin, D.~L. 2013.
\newblock Repairing trust with individuals vs. groups.
\newblock \emph{Organizational Behavior and Human Decision Processes}, 120(1):
  1--14.

\bibitem[{Kittmann et~al.(2015)Kittmann, Fröhlich, Schäfer, Reiser,
  Weißhardt, and Haug}]{kittmann_let_2015}
Kittmann, R.; Fröhlich, T.; Schäfer, J.; Reiser, U.; Weißhardt, F.; and
  Haug, A. 2015.
\newblock Let me {Introduce} {Myself}: {I} am {Care}-{O}-bot 4, a {Gentleman}
  {Robot}.
\newblock In \emph{Let me {Introduce} {Myself}: {I} am {Care}-{O}-bot 4, a
  {Gentleman} {Robot}}, 223--232. De Gruyter.
\newblock ISBN 978-3-11-044392-9.

\bibitem[{Knox and Stone(2009)}]{knox2009interactively}
Knox, W.~B.; and Stone, P. 2009.
\newblock Interactively shaping agents via human reinforcement: The TAMER
  framework.
\newblock In \emph{Proceedings of the fifth international conference on
  Knowledge capture}, 9--16.

\bibitem[{Kohn et~al.(2021)Kohn, de~Visser, Wiese, Lee, and
  Shaw}]{kohn_measurement_2021}
Kohn, S.~C.; de~Visser, E.~J.; Wiese, E.; Lee, Y.-C.; and Shaw, T.~H. 2021.
\newblock Measurement of {Trust} in {Automation}: {A} {Narrative} {Review} and
  {Reference} {Guide}.
\newblock \emph{Frontiers in Psychology}, 12.

\bibitem[{Kostavelis et~al.(2016)Kostavelis, Giakoumis, Malasiotis, and
  Tzovaras}]{kostavelis_ramcip_2016}
Kostavelis, I.; Giakoumis, D.; Malasiotis, S.; and Tzovaras, D. 2016.
\newblock {RAMCIP}: {Towards} a {Robotic} {Assistant} to {Support} {Elderly}
  with {Mild} {Cognitive} {Impairments} at {Home}.
\newblock In Serino, S.; Matic, A.; Giakoumis, D.; Lopez, G.; and Cipresso, P.,
  eds., \emph{Pervasive {Computing} {Paradigms} for {Mental} {Health}},
  Communications in {Computer} and {Information} {Science}, 186--195. Cham:
  Springer International Publishing.
\newblock ISBN 978-3-319-32270-4.

\bibitem[{Langer et~al.(2019)Langer, Feingold-Polak, Mueller, Kellmeyer, and
  Levy-Tzedek}]{langer_trust_2019}
Langer, A.; Feingold-Polak, R.; Mueller, O.; Kellmeyer, P.; and Levy-Tzedek, S.
  2019.
\newblock Trust in socially assistive robots: {Considerations} for use in
  rehabilitation.
\newblock \emph{Neuroscience \& Biobehavioral Reviews}, 104: 231--239.

\bibitem[{Lee and Riek(2018)}]{lee_reframing_2018}
Lee, H.~R.; and Riek, L.~D. 2018.
\newblock Reframing {Assistive} {Robots} to {Promote} {Successful} {Aging}.
\newblock \emph{ACM Transactions on Human-Robot Interaction}, 7(1):
  11:1--11:23.

\bibitem[{Malle and Ullman(2021)}]{malle_multidimensional_2021}
Malle, B.~F.; and Ullman, D. 2021.
\newblock A multidimensional conception and measure of human-robot trust.
\newblock In \emph{Trust in {Human}-{Robot} {Interaction}}, 3--25. Elsevier.
\newblock ISBN 978-0-12-819472-0.

\bibitem[{Martinez-Martin, Costa, and
  Cazorla(2019)}]{martinez-martin_pharos_2019}
Martinez-Martin, E.; Costa, A.; and Cazorla, M. 2019.
\newblock {PHAROS} 2.0—{A} {PHysical} {Assistant} {RObot} {System}
  {Improved}.
\newblock \emph{Sensors (Basel, Switzerland)}, 19(20): 4531.

\bibitem[{Merritt et~al.(2013)Merritt, Heimbaugh, LaChapell, and
  Lee}]{merritt_i_2013}
Merritt, S.~M.; Heimbaugh, H.; LaChapell, J.; and Lee, D. 2013.
\newblock I {Trust} {It}, but {I} {Don}’t {Know} {Why}: {Effects} of
  {Implicit} {Attitudes} {Toward} {Automation} on {Trust} in an {Automated}
  {System}.
\newblock \emph{Human Factors}, 55(3): 520--534.
\newblock Publisher: SAGE Publications Inc.

\bibitem[{Mišeikis et~al.(2020)Mišeikis, Caroni, Duchamp, Gasser, Marko,
  Mišeikienė, Zwilling, de~Castelbajac, Eicher, Früh, and
  Früh}]{miseikis_lio-personal_2020}
Mišeikis, J.; Caroni, P.; Duchamp, P.; Gasser, A.; Marko, R.; Mišeikienė,
  N.; Zwilling, F.; de~Castelbajac, C.; Eicher, L.; Früh, M.; and Früh, H.
  2020.
\newblock Lio-{A} {Personal} {Robot} {Assistant} for {Human}-{Robot}
  {Interaction} and {Care} {Applications}.
\newblock \emph{IEEE Robotics and Automation Letters}, 5(4): 5339--5346.
\newblock Conference Name: IEEE Robotics and Automation Letters.

\bibitem[{Natarajan and Gombolay(2020)}]{3374839}
Natarajan, M.; and Gombolay, M. 2020.
\newblock Effects of Anthropomorphism and Accountability on Trust in Human
  Robot Interaction.
\newblock In \emph{Proceedings of the 2020 ACM/IEEE International Conference on
  Human-Robot Interaction}, HRI '20, 33–42. New York, NY, USA: Association
  for Computing Machinery.
\newblock ISBN 9781450367462.

\bibitem[{{Paro Robotics}(1998)}]{paro}
{Paro Robotics}. 1998.
\newblock {PARO Therapeutic Robot}.
\newblock Http://www.parorobots.com/.

\bibitem[{Raub(1981)}]{raub_correlates_nodate}
Raub, A.~C. 1981.
\newblock Correlates of {Computer} {Anxiety} in {College} {Students}.

\bibitem[{Robinette, Howard, and Wagner(2015)}]{robinette_timing_2015}
Robinette, P.; Howard, A.; and Wagner, A. 2015.
\newblock \emph{Timing is {Key} for {Robot} {Trust} {Repair}}.

\bibitem[{{SoftBank Robotics}(2008)}]{nao}
{SoftBank Robotics}. 2008.
\newblock {NAO} the humanoid and programmable robot.
\newblock Https://www.softbankrobotics.com/emea/en/nao.

\bibitem[{{Softbank Robotics}(2014)}]{pepper}
{Softbank Robotics}. 2014.
\newblock Pepper the humanoid and programmable robot.
\newblock Https://www.softbankrobotics.com/emea/en/pepper.

\bibitem[{Ullman and Malle(2018)}]{ullman_what_2018}
Ullman, D.; and Malle, B.~F. 2018.
\newblock What {Does} it {Mean} to {Trust} a {Robot}? {Steps} {Toward} a
  {Multidimensional} {Measure} of {Trust}.
\newblock In \emph{Companion of the 2018 {ACM}/{IEEE} {International}
  {Conference} on {Human}-{Robot} {Interaction}}, {HRI} '18, 263--264. New
  York, NY, USA: Association for Computing Machinery.
\newblock ISBN 978-1-4503-5615-2.

\bibitem[{Umbrico et~al.(2020)Umbrico, Cesta, Cortellessa, and
  Orlandini}]{umbrico_holistic_2020}
Umbrico, A.; Cesta, A.; Cortellessa, G.; and Orlandini, A. 2020.
\newblock A {Holistic} {Approach} to {Behavior} {Adaptation} for {Socially}
  {Assistive} {Robots}.
\newblock \emph{International Journal of Social Robotics}, 12(3): 617--637.

\bibitem[{van Wynsberghe(2013)}]{van_wynsberghe_designing_2013}
van Wynsberghe, A. 2013.
\newblock Designing robots for care: care centered value-sensitive design.
\newblock \emph{Science and Engineering Ethics}, 19(2): 407--433.

\bibitem[{Vitanza et~al.(2019)Vitanza, D’Onofrio, Ricciardi, Sancarlo, Greco,
  and Giuliani}]{vitanza_assistive_2019}
Vitanza, A.; D’Onofrio, G.; Ricciardi, F.; Sancarlo, D.; Greco, A.; and
  Giuliani, F. 2019.
\newblock Assistive {Robots} for the {Elderly}: {Innovative} {Tools} to
  {Gather} {Health} {Relevant} {Data}.
\newblock In Consoli, S.; Reforgiato~Recupero, D.; and Petković, M., eds.,
  \emph{Data {Science} for {Healthcare}: {Methodologies} and {Applications}},
  195--215. Cham: Springer International Publishing.
\newblock ISBN 978-3-030-05249-2.

\bibitem[{Yagoda and Gillan(2012)}]{yagoda_you_2012}
Yagoda, R.~E.; and Gillan, D.~J. 2012.
\newblock You {Want} {Me} to {Trust} a {ROBOT}? {The} {Development} of a
  {Human}–{Robot} {Interaction} {Trust} {Scale}.
\newblock \emph{International Journal of Social Robotics}, 4(3): 235--248.

\bibitem[{Šabanović et~al.(2015)Šabanović, Chang, Bennett, Piatt, and
  Hakken}]{sabanovic_robot_2015}
Šabanović, S.; Chang, W.-L.; Bennett, C.~C.; Piatt, J.~A.; and Hakken, D.
  2015.
\newblock A {Robot} of {My} {Own}: {Participatory} {Design} of {Socially}
  {Assistive} {Robots} for {Independently} {Living} {Older} {Adults}
  {Diagnosed} with {Depression}.
\newblock In Zhou, J.; and Salvendy, G., eds., \emph{Human {Aspects} of {IT}
  for the {Aged} {Population}. {Design} for {Aging}}, Lecture {Notes} in
  {Computer} {Science}, 104--114. Cham: Springer International Publishing.
\newblock ISBN 978-3-319-20892-3.

\end{thebibliography}

\end{document}